\documentclass{edm_article}

\usepackage{graphicx}
\usepackage{algorithm}
\usepackage{booktabs}
\usepackage{algorithmic}
\usepackage{tcolorbox}
\usepackage{array}
\usepackage{newfloat}
\usepackage{listings}
\usepackage{url}
\usepackage{float}
\usepackage{placeins} 
\usepackage{subcaption}
\usepackage[numbers]{natbib}

\begin{document}


\title{Using Large Multimodal Models to Extract Knowledge Components for Knowledge Tracing from Multimedia Question Information}


\numberofauthors{4}
\author{
  \alignauthor Hyeongdon Moon\\
         \affaddr{EPFL}\\         \email{donim@andrew.cmu.edu}
\and
  \alignauthor Richard Lee Davis\\
         \affaddr{KTH Royal Institute of Technology}\\
         \email{rldavis@kth.se}
\and
  \alignauthor Seyed Parsa Neshaei\\
         \affaddr{EPFL}\\         \email{seyed.neshaei@epfl.ch}
\and
  \alignauthor Pierre Dillenbourg\\
         \affaddr{EPFL}\\         \email{pierre.dillenbourg@epfl.ch}
}

\maketitle


\begin{abstract}
Knowledge tracing models have enabled a range of intelligent tutoring systems to provide feedback to students.
However, existing methods for knowledge tracing in learning sciences are predominantly reliant on statistical data and instructor-defined knowledge components, making it challenging to integrate AI-generated educational content with traditional established methods. We propose a method for automatically extracting knowledge components from educational content using instruction-tuned large multimodal models. We validate this approach by comprehensively evaluating it against knowledge tracing benchmarks in five domains. Our results indicate that the automatically extracted knowledge components can effectively replace human-tagged labels, offering a promising direction for enhancing intelligent tutoring systems in limited-data scenarios, achieving more explainable assessments in educational settings, and laying the groundwork for automated assessment. \footnote{Codes are available at \url{https://github.com/DoniMoon/LLMKT}}
\end{abstract}



\section{Introduction}
\label{sec:intro}

\begin{figure}[ht]
    \Description{A flowchart illustrating the four main steps of the experiment. From top to bottom: “Problem Content” feeds into “LMM Inference,”
    “Knowledge Descriptions” feeds into “Calculate Sentence Embedding,”
    “Embedding Vectors” feeds into “Clustering,” and
    “Knowledge Components” feed into “Evaluate.” The final step, “Evaluate,” references two validation tasks: Additive Factor Model (AFM) and Knowledge Tracing (KT), which are linked to Tables 3 and 4 in the paper.}
    \centering
    \includegraphics[width=0.9\linewidth]{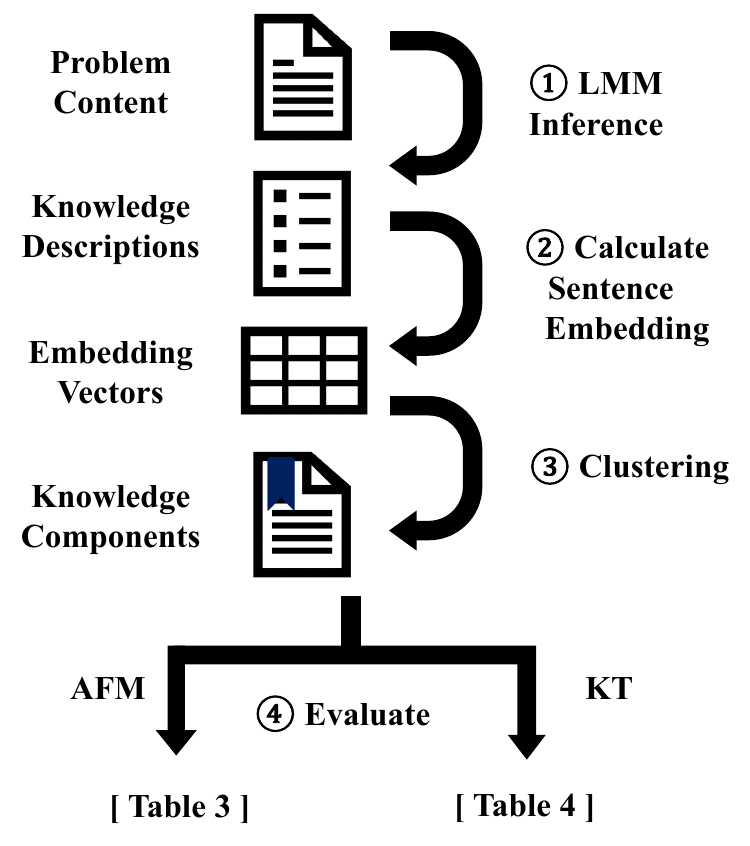}
    \caption{Overview of our experiment. We evaluate the quality of extracted knowledge components by utilizing them in two validation tasks, Knowledge Tracing (KT) and Additive Factor Model (AFM).}
    \label{fig:traditional}
\end{figure}

Intelligent Tutoring Systems (ITS) are advanced computer programs that provide personalized and adaptive educational instruction to learners. For over half a century, they have been a subject of active research and discussion in the interdisciplinary field of education and artificial intelligence \cite{mousavinasab2021intelligent}. These systems integrate techniques from artificial intelligence to deliver tailored instruction, dynamically adjusting to the needs and progress of individual students. By simulating one-on-one tutoring, ITS offer immediate and specific feedback, enhancing student engagement and improving learning outcomes. This combination of real-time adaptability and personalized feedback makes ITS a valuable tool in modern education, bridging the gap between traditional classroom methods and individualized learning approaches \cite{nwana1990intelligent}.

Knowledge Tracing (KT) is a foundational task in Intelligent Tutoring Systems (ITS), aiming to model a student’s knowledge state and predict future performance on educational tasks. Traditional KT models rely heavily on statistical techniques to analyze historical problem-solving data. Early approaches, such as Bayesian Knowledge Tracing (BKT), used hidden Markov models to estimate the probability of a student knowing a particular skill at any given time \cite{corbett1994knowledge}.

More recent models have leveraged advancements in machine learning, particularly deep learning, to enhance predictive accuracy \cite{abdelrahman2023knowledge}. Deep Knowledge Tracing (DKT) was one of the first models to apply recurrent neural networks (RNNs) to KT, demonstrating significant improvements over BKT by capturing the sequential nature of students' learning processes. Subsequent models such as Self-Attentive Knowledge Tracing (SAKT) and Separated Self-AttentIve Neural Knowledge Tracing (SAINT) have further refined these approaches \cite{choi2020towards, pandey2019self}.

Knowledge Components (KCs) are defined within the Knowledge-Learning-Instruction framework as \textit{acquired units of cognitive function or structure that can be inferred from performance on a set of related tasks} \cite{koedinger2012knowledge}. While we cannot directly observe the changes in a student's KCs, we infer them through interactions during assessment and instruction. In Knowledge Tracing, each problem is often labeled with its corresponding KCs, which are typically assigned by human experts based on their expected relevance to the problem and statistically validated for their ability to explain student performance.

There are several methods to define these KCs. \cite{pelanek2017bayesian} introduces four categories of domain modeling: `KCs as disjoint sets', `Multiple KCs per item', `Hierarchy of KCs', and `KCs with prerequisites'. In the KT domain, the most common approaches are modeling KCs as disjoint sets or mapping multiple KCs to an item. The simplest approach is to map each question to a single KC, as used in Bayesian KT, SAKT, and SAINT. When mapping multiple KCs, the relationship between KCs and items is represented in a Q-matrix, where rows represent questions and columns represent KCs. Cognitive Diagnosis Models (CDM) like DINA, NIDA, and generalized DINA use this Q-matrix for KT \cite{de2011generalized}, thus supporting multiple KCs labeled for each question.

\begin{figure*}[ht!]
    \Description{A schematic flow for extracting and grouping knowledge components from a sample question about force direction on a cable. On the left, a monitor labeled “GPT-4o” displays the question “What is the direction and sense of the force exerted on the displayed section of the cable?” Below it, bullet points summarize extracted knowledge components (e.g., identifying force directions, understanding tension in cables, and analyzing pulleys). On the right, these components are grouped under two headings: “Moment calculation” (focusing on coordinate points and torque) and “Trigonometric functions” (focusing on formulas for tension). Colored stacks of bars illustrate clustering of the knowledge components based on sentence embeddings, forming semantically related groups.}
    \centering
    \includegraphics[width=0.9\linewidth]{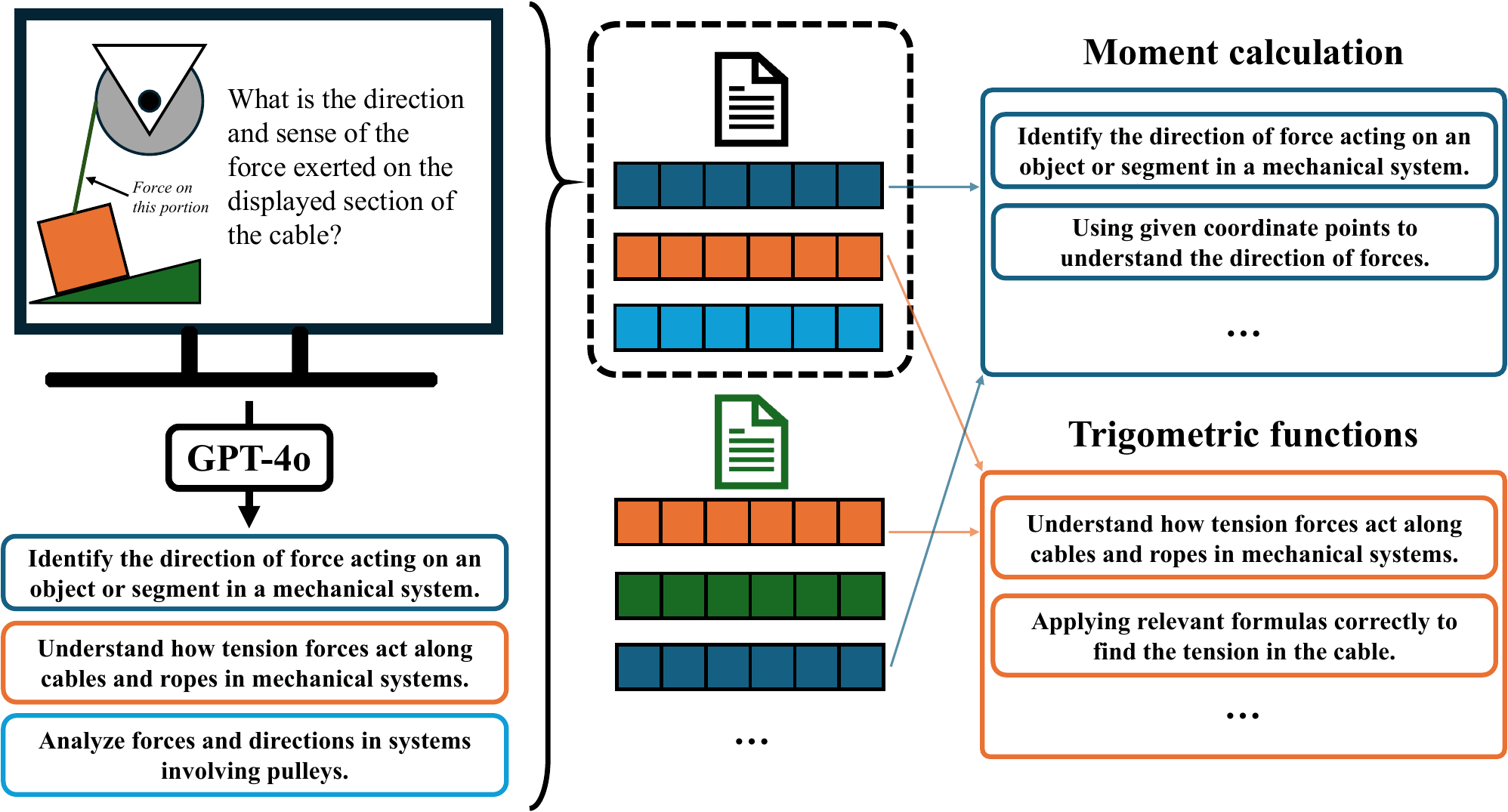}
    \caption{Overall procedure of extracting knowledge components. First, we extract knowledge components using GPT-4o API from the parsed question content. Then, we calculate sentence embedding for all descriptions of the generated knowledge components. At last, we group each knowledge components into similar semantic group using a clustering approach.}
    \label{fig:procedure}
\end{figure*}


Well-defined KCs are essential for accurately predicting learner performance, building effective adaptive learning systems, and providing efficient learning support and improvement. Due to this importance, research has been conducted to improve KC models through Difficulty Factors Assessment (DFA) and to evaluate the improvements using the Additive Factors Model (AFM), ultimately creating better learner models \cite{liu2017going,koedinger2010data}.

Our research proposes a novel approach leveraging the instruction-tuned Large Multimodal Model (LMM) to extract and utilize Knowledge Components (KCs) from educational content. Figure \ref{fig:traditional} illustrates the overall architecture of our approach.
Our method involves parsing educational materials to extract text and images, using the GPT-4o API to identify and describe inherent KCs, and clustering similar components based on sentence embeddings. This automation not only improves KC extraction but also enhances the prediction of student performance on new and unseen content.

We observed performance improvements when using LMM-generated KCs as additional features in various KT methods. For example, in the Performance Factors Analysis (PFA) method, LMM-generated KCs resulted in greater performance increases compared to human-generated KCs. Other KT methods also showed comparable performance improvements. Additionally, we analyzed the performance factors of generated KCs using the AFM. 

We compared LMM-generated KCs to human-generated KCs by using them in four different knowledge tracing models across six datasets.
Overall, when using LMM-generated KCs, we demonstrated comparable or superior performance to human-generated KCs.
To encourage further research in this area, we have refined and publicly released KT benchmarks with content data across five domains. By providing these benchmarks, we aim to facilitate the development of more advanced KT methodologies that can fully utilize the potential of LMM-generated KCs.

In summary, our contributions are three-fold:
\begin{enumerate}
    \item Introducing a novel zero-shot KC generation methodology that can be applied to general domains and diverse modalities supported by LMMs.
    \item Our automatically generated KCs model students' prob-lem-solving data as effectively as human-created KCs in both KT and AFM.
    \item We publish a reproducible KT benchmark with parsed content data, advancing content-aware Knowledge Tracing methods. 
\end{enumerate}

\section{Related Work}

\subsection{Utilizing LLMs to improve ITS and Knowledge Tracing}

The emergence of advanced NLP tools in recent years, especially instruction-tuned LLMs like ChatGPT, has significantly enhanced ITS by providing natural, human-like interactions \cite{limo2023personalized, grassini2023shaping}.
Large language models (LLMs) have been shown to provide support in a range of areas, including, but not limited to, planning learning instruction \cite{hu2024teaching}, scaffolding \cite{goslen2024llm, liu2024scaffolding}, and helping students solve math word problems \cite{he2023solving}.
They assist learners in multiple ways, such as participating in back-and-forth instructional conversations \cite{lieb2024student, salminen2024using} or providing feedback to students \cite{dai2023can, gubelmann2024exploring}.

Moreover, LLMs have also been used to improve the accuracy and performance of the KT models in the backbone of ITS. LM-KT was proposed to perform KT even when there is no prior problem-solving data from the student. The LM-KT model trains GPT-2 to perform Knowledge Tracing on content without any prior student interaction records \cite{srivastava2021question}. In a basic second-language acquisition problem space, LM-KT models student success rates based on the input of natural language sentences as questions. 
Following LM-KT, other work has shown that leveraging the generalizability of LLMs can enhance the performance of Knowledge Tracing \cite{neshaei2024towards, 10.1007/978-3-031-64302-6_13} and address cold-start problems \cite{lee2024language}. However, their applicability to other domains is limited, as the language model needs to be fine-tuned on the specific domain.


LLMs have the ability to digest information from source documents which can help with extracting information from large-scale educational data; for example, they have been used for summarize content by intention \cite{shin2022dialogue} or extracting key points out of educational data \cite{dagdelen2024structured}. However, the application of LLMs to improve the performance of KT models, specifically by assisting in the process of KC extraction, has been rare in the literature. We address this gap by providing and evaluating a structured KC extraction method using LLMs in the loop, with the goal of improving the accuracy of KC models while reducing the need for human labeling labor.

\subsection{Extracting Knowledge Components}

Commonly, the assignment of KCs to questions is done by human experts, e.g., instructors \cite{8295250}; however, this requires manual labor, making it less suitable for course offerings with a high number of questions. As a result, previous researchers have explored methods to move from expert-annotated KCs towards extracting KCs from the problem information automatically, which can aid in developing more accurate KT models. For example, a methodology to extract KCs from documents using classical NLP techniques and annotate these documents for application in adaptive online textbooks has been proposed \cite{thaker2019student}. Additionally, there is a study that validated the accuracy of a method for automatically generating skills for problems by fine-tuning a language model on problem and skill label data to enable computer adaptive tests \cite{8295250}.

Unlike previous studies, we perform the task of tagging the Knowledge Components (KCs) required by questions without using pre-trained problem data or provided schemas of knowledge components. Additionally, we focus on generating entities that clearly correspond to the previous concept of `knowledge component', rather than simply using terms like skill, knowledge, or tag, and we conduct validation for the KCs. The study most closely related to ours analyzed the relation of questions and KCs using LLMs \cite{moon2022evaluating}. This study uses pre-trained LLMs to verify that if a problem generated from a single KC is a good one, there is a strong dependency on that KC, for the purpose of evaluating automatically generated questions. Conversely, we aim to extract KCs directly from the questions, under the assumption that each question is designed to test specific knowledge.




\section{Methodology}
\label{sec:methodology}
As illustrated in Figure \ref{fig:procedure}, our overall pipeline involves preprocessing student interaction data enriched with content information, extracting KCs, evaluating the quality of the extracted KCs, and analyzing their utility across various KT methods.

\begin{figure*}[ht]
    \Description{Two line charts display clustering metrics for varying numbers of clusters (from 0 to 200) in the “oli_statics” dataset. The left chart shows the Elbow Method, with the y-axis labeled “WCSS,” which decreases as the number of clusters increases, forming a downward-sloping curve. The right chart shows the Silhouette Score, which rises gradually on the y-axis as the number of clusters increases, indicating how well-separated the clusters are.}
    \centering
    \includegraphics[width=\textwidth, trim= 0cm 0cm 0cm 1.1cm, clip]{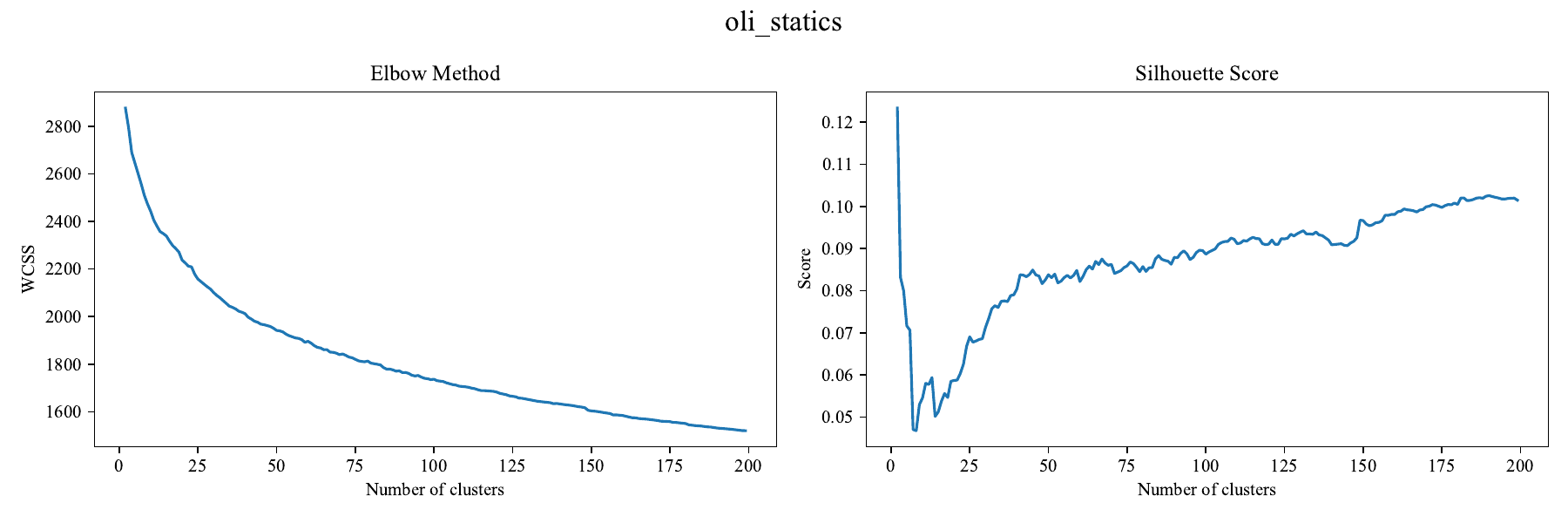}   
    \caption{Clustering Scores used to determine the number of clusters in oli\_statics. The left graph (Elbow Method) shows the WCSS (y-axis) values for different numbers of clusters (x-axis), indicating that WCSS gradually decreases as the number of clusters increases. The right graph (Silhouette Score) presents the silhouette scores measured over the same range of cluster numbers, generally showing that the score tends to rise and fluctuate as the number of clusters increases.}
    \label{fig:statics_cluster}
\end{figure*}

\subsection{Dataset}

We chose to process the OLI datasets from CMU Datashop\footnote{https://pslcdatashop.web.cmu.edu/} since the OLI\_Statics2011 dataset is a well-known benchmark for knowledge tracing which has publicly available content data\cite{koedinger2010data, abdelrahman2023knowledge}. The Open Learning Initiative (OLI) project provides research-based courseware suitable for various class formats and supports advanced research. We gained access to several domains of OLI learning content from CMU Datashop. The preprocessing code for the datasets is publicly available, and to facilitate reproduction while protecting the content data, the parsed files for each dataset have been uploaded to the corresponding entries in Datashop.

From CMU Datashop, all transaction data was extracted by selecting “all data” in the “Export” tab and choosing “By Transaction” in the detailed options. The content data was collectively downloaded by clicking the “Download Problem Content” button in the “Dataset Info/Problem List” tab. We used the swftools package \footnote{http://www.swftools.org/} in the Windows OS environment to extract images from the SWF file, which is not supported anymore due to the deprecation of Flash support. 

all mp3 files are converted into text files, using whisper-large-v2 \cite{radford2022whisper}. 

\subsubsection{OLI Engineering Statics}

For the Statics dataset, we used the Fall 2011 version\footnote{\url{https://pslcdatashop.web.cmu.edu/DatasetInfo?datasetId=507}}. While other versions included a wider variety of KC models, the number of users and transactions was smaller compared to other statics benchmarks. Therefore, we considered the Fall 2011 version to be the superset and used it. The KC model named F2011 was used as the human tag. The original 361,092 transactions were reduced by about half to 189,047.

\subsubsection{OLI Principles of Computing}

The content of the OLI computing dataset \footnote{\url{https://pslcdatashop.web.cmu.edu/DatasetInfo?datasetId=1806}} was different from other subjects, as it contained content from multiple subjects under the root directory. To maintain consistency with the method used for processing other datasets, only the content from Principles\_of\_Computing\_v\_1\_10, which includes the version in the subject name, was used. The KC models \texttt{principles\_of\_computing\_1\_13} and \texttt{principles\_of\_computing\_1\_10} were used to calculate AFM scores. The transaction data was filtered to include only those containing this content. As a result, the original transaction dataset, which had 37,233 rows, was reduced to 16,951 rows.

\subsubsection{OLI French}

We used the `French1 - Spring 2014' dataset for the French dataset\footnote{\url{https://pslcdatashop.web.cmu.edu/DatasetInfo?datasetId=918}}. Since it included various KC models, we used \textit{Bonnie’s Model}, which had the best AFM performance. To train AFM, \textit{Level4} KC model, which also reports good AFM performance, is used. This dataset contained many items with voice mp3 files, so we converted these files to text using the whisper-large-v2 model and inserted the transcriptions into the question text. At the position in the HTML where the mp3 execution was embedded, we prefixed the converted text with `[transcription of embedded mp3 file]:'. Out of the total 278,489 rows of transactions, the final remaining data consisted of 53,255 rows.

\subsubsection{OLI Biology}

We used the `Oli\_biology' dataset from the `Bridge to Success' project for the Biology data\footnote{\url{https://pslcdatashop.web.cmu.edu/DatasetInfo?datasetId=1148}}. The KC model named \textit{intro\_biology-1\_0} was used as the human label. After processing, 3,285,685 rows remained out of the original 5,852,795 rows of transactions.

\subsubsection{OLI Psychology}

We used the `Psychology MOOC GT - Spring 2013' dataset for the Psychology data\footnote{\url{https://pslcdatashop.web.cmu.edu/DatasetInfo?datasetId=863}} and utilized the KC model named \textit{psychology-1-4}. After processing, 1,935,496 rows remained out of the original 2,493,609 rows of transactions.

\begin{figure*}[ht]
\Description{Two boxes list example knowledge components from a statics dataset. The upper box, labeled “Trigonometric Relationships,” contains items about decomposing forces into perpendicular components and applying trigonometric functions to resolve vectors. The lower box, labeled “Moment Calculation,” includes items about determining force direction and sense, recognizing how forces contribute to equilibrium, and computing the moment of a force. Each bullet point corresponds to a specific sub-topic within these clusters.}
\centering
  \begin{tcolorbox}
    \small
    Understanding how to calculate the x and y components of forces. \\
    Applying trigonometric functions to resolve forces into perpendicular components. \\
    Understand how to decompose a force into its perpendicular components. \\
    Calculate the position vector from point O to the point of application of each force. \\
    Learn how to represent forces using vectors.\\
    Decomposing a given force into its x and y components based on angles provided. \\
    Decompose a force into its components.\\
    Using trigonometric functions to resolve forces into components.\\
    Calculate the x and y components of a vector given its magnitude and direction. \\
    Use sine and cosine functions to resolve forces into their components.
  \end{tcolorbox}

  \begin{tcolorbox}[colback=green!5!white]
    \small
    Understand how to determine sense and direction of a force. \\
    Assigning labels to the identified forces based on their origin and point of interaction. \\
    The direction and point of application of a force determine how it contributes to the equilibrium conditions. \\
    Recognize that forces can act in multiple directions. \\
    Understanding the appropriate direction of force at a given point. \\
    Determine the resultant direction of the applied force from one body to another. \\
    Determine the force labels in the context of the question. \\
    Understanding the direction and magnitude of vertical forces in a system. \\
    The vector represents the magnitude and direction of a force. \\
    Predicting the directions of forces exerted at a joint based on given resultant forces.
  \end{tcolorbox}

  \caption{Example of the knowledge components belonging to the cluster \textbf{Trigonometric Relationships} (above) and \textbf{Moment Calculation} (below) in \textit{Statics} dataset.}
  \label{clustered_example}
\end{figure*}

\subsection{Knowledge Component Extraction}
To leverage the capabilities of instruction-tuned LLMs for educational purposes, we first focus on extracting knowledge components from educational content. Our approach involves the following steps:
\subsubsection{Content Processing}
The content data available from the CMU DataShop consists of HTML pages representing the learning materials that students interact with \cite{koedinger2010data}. We parsed this data, extracting text and images from Flash files embedded within the content, and converted MP3 files into text data using the \textit{whisper-large-v2} model \cite{radford2022whisper}. We processed the image files to embed them in the Chat Template for the OpenAI API \footnote{https://platform.openai.com/docs/guides/vision}, preserving their positions and order from the original HTML content where the images were located. Each problem was matched with the corresponding steps in the CMU DataShop transaction data, creating datasets for five subjects: Statics, Psychology, Biology, Computing, and French.

\subsubsection{Knowledge Component Extraction Prompt}
Using the OpenAI API for the GPT-4o model, we extracted the knowledge components for each problem as shown in Figure \ref{fig:procedure}. Each problem can have multiple knowledge components, with each component consisting of a name field (1–3 words) and a description field (1–2 sentences). We applied this basic zero-shot prompt from Fig.~\ref{fig:procedure} only once and, being satisfied with the qualitative output, did not undertake any further prompt tuning.

\subsubsection{Clustering Knowledge Components}
The extracted KCs consisted of natural language sentences, which occasionally referred to the same topic but with different sets of words. To utilize these components for problem correlation and be able to assign the same identifiers to semantically similar KCs, we needed to group them together. To do this, we computed sentence embeddings for each component and performed clustering based on similarity. The optimal number of clusters was determined by maximizing the silhouette score of the clustering \cite{rousseeuw1987silhouettes}. We compared the performance of the Sentence-T5-XXL model \cite{ni-etal-2022-sentence} and OpenAI's \textit{text-embedding-3-large} model for this task \footnote{\url{https://openai.com/index/new-embedding-models-and-api-updates/}}.

Figure \ref{fig:statics_cluster} shows the WCSS and Silhouette Score of the K-means clustering method as the number of clusters varies from 2 to 200 in the oli\_statics dataset. Due to the instability of the Silhouette Score when the number of clusters is very small, we analyzed cases where the number of clusters is greater than 10. Each point in Figure \ref{fig:cluster_score} represents the local maximum Silhouette Score within one of the 10 bins divided between the range of 10 to 200 clusters, and the AFM performance was measured using these cluster numbers.

Meanwhile, for zero-shot implementation, using more than 100 KCs, which maximizes the Silhouette Score, made it impossible to have at least one problem with each KC in both the train and test splits. Therefore, we used the number of clusters at the local maximum of the third bin, where such splits were feasible for all datasets. The selected numbers of clusters were 52, 63, 52, 61, and 49 for computing, statics, French, psychology, and biology, respectively.

\subsection{Knowledge Component Quality Evaluation}
\label{section:kc_quality}
To validate the effectiveness of our knowledge component extraction method, we conducted a comprehensive quality evaluation across five different datasets. Using the Additive Factors Model (AFM), we measured the Root Mean Square Error (RMSE) and compared it with the RMSE of human-generated KC mappings for each dataset. This evaluation provided a robust assessment of the accuracy and reliability of the LLM-extracted knowledge components.

The KCs used in this validation were evaluated by measuring the silhouette score of K-means clustering from 2 to 200 clusters. We automatically determined the optimal number of clusters by detecting the knee point of the silhouette score change \cite{rousseeuw1987silhouettes, satopaa2011finding}. To further verify the quality of KCs based on the level of clustering, we divided the entire range of 2 to 200 clusters into ten sections. For each section, we identified the point where the silhouette score was at its local maximum and observed the performance change of the AFM when KCs were generated with the corresponding number of clusters.

RMSE was measured not only for the entire dataset but also in environments where student ID information and item ID information were masked, respectively. This ablation experiment was done to identify which biases the model was utilized to demonstrate performance. For performance measurement, we utilized PyAFM, a Python implementation of the AFM \cite{afmslip:2015}.

To better understand the Knowledge Components (KCs) categorized through clustering, the most frequent name within each cluster was selected as the representative name for that cluster, alongside the descriptions included in that cluster. As illustrated in Figure \ref{clustered_example}, these names typically consist of 2-3 words, such as \textbf{Trigonometric Relationships} or \textbf{Moment Calculation}. In most clusters, approximately one-third of the items shared the same name, which was then chosen as the mode and designated as the cluster’s representative name.

\subsection{Knowledge Tracing Baselines Using KCs}
\label{sec:kc_baselines}
Using the extracted knowledge components, we used code from prior work \cite{gervet2020deep} to conduct KT with IRT, PFA, DAS3H, SAKT, and DKT. In our implementation of PFA, we also considered incorporating a student-specific intercept—an enhancement reported to boost performance in recent work \cite{chu2023predictiveness}—but observed degraded results across all configurations; consequently, we omitted student intercept encoding in our final PFA setup.

In the experiments of the previous section, the performance of the OpenAI embedding model was superior to that of the T5-XXL model, so we evaluated the KCs generated using the OpenAI \textit{text-embedding-3-large} model. Among these, IRT is an algorithm that does not use KCs, while PFA and DAS3H use a Q-matrix. SAKT and DKT operate in environments with disjoint KCs, so we used joint skill assignments. Joint skill considers the combination of KCs assigned to each question as a single KC \cite{xiong2016going}. For example, a problem with KC 1 and KC 2 is treated as having a distinct KC different from a problem with only KC 1.

To compare the impact of the KCs generated by our algorithm, we prepared four types of baselines. The \textbf{Random} setting assigns all KCs randomly as a baseline. The \textbf{Human} setting uses the highest-performing human-generated KCs tagged in each dataset from CMU Datashop.

In addition, we also measured the zero-shot KT performance on completely unseen items. To achieve this, we created a train-test split where no items overlap between the train and test sets, but the KCs of the items in the test set appear at least once in the train set. Then, we applied logistic regression to the KCs-aware features to determine the zero-shot performance across each domain. By splitting the dataset in this manner, we ensured that the model's ability to generalize to new items was rigorously tested. This approach allowed us to evaluate the robustness and adaptability of our LLM-generated KCs in handling novel content.

\subsection{Detailed Experiment Settings}
\label{app:detailed_setting}

\subsubsection{LMM Inference}
\begin{figure}[ht]
\Description{A rectangular prompt box labeled “Prompt” displays instructions for extracting knowledge components needed to solve a question. It specifies that each component should have a short name (2–4 words) and a single-sentence description, and the output should be in JSON format. An example JSON snippet is shown, containing one knowledge component with a name and description.}
    \centering
\begin{tcolorbox}[colframe=blue!75!black, colback=blue!10, title=Prompt]
Extract the knowledge components required to solve this question. Each knowledge component has two fields:
\begin{itemize}
    \item Name: 2 to 4 words
    \item Description: 1 sentence
\end{itemize}
Output is in JSON format, like:
\begin{verbatim}
{
  "knowledge_components": [
    {
      "name": "knowledge component 1",
      "description": "understand
      how to apply kc 1"
    }
  ]
}
\end{verbatim}
\end{tcolorbox}
    \caption{GPT-4o prompt used for knowledge components extraction}
    \label{app:prompt}
\end{figure}

Figure \ref{app:prompt} shows the prompt used as the system role. The random seed for the API was set to 42, and the problem was provided in the user role. The problem was appended with the postfix ‘Content:\textbackslash n\textbackslash n’. Images were uploaded and encoded in base64, and the prompt was generated to place the images between the text, preserving their positions as closely as possible to their actual locations in the problem.

\subsubsection{Zero-Shot Knowledge Tracing}
\label{appendix:zero-shot}
We implemented zero-shot knowledge tracing using the same codebase\footnote{\url{https://github.com/theophilegervet/learner-performance-prediction}} that was used for KT implementation\cite{gervet2020deep}. The logistic regression setup was configured to experiment with various combinations of features, specifically utilizing the s, sc, tc, tw, w, and a tags. These options allow us to use features that record which KCs are present, how many times each KC appears in the user’s history, the total number of problems the user has solved, the total number of problems the user has correctly solved, and how many attempts were made for each problem within a time window. When selecting these options, we understood the code and chose all relevant options, but did not compare with other options due to the absence of a validation set.

\subsection{Quality of the Generated Knowledge Components}

\begin{table*}[h]
\centering
\caption{Examples of assigned Knowledge Components and the descriptions}
\begin{subtable}{\textwidth}
\centering

\caption{oli\_statics}
\begin{tabular}{|p{3.7cm}|p{7cm}|p{3.7cm}|}
\hline
\textbf{Generated Name} & \textbf{Description} & \textbf{KC Name} \\ \hline

\multicolumn{3}{|l|}{\textbf{Item ID: 1273}} \\ \hline
Reading comprehension & Understand the text of the question and the options provided. & identifying correct option \\
Multiple choice format & Recognize the structure of a multiple choice question and how to select an answer. & understanding question format \\
Decision making & Decide between the given options based on the question's requirements. & identifying correct option \\ \hline

\multicolumn{3}{|l|}{\textbf{Item ID: 1097}} \\ \hline
Summation of Forces & Understand that $\Sigma$F y = 0 denotes the summation of all forces in the y-direction equaling zero. & Force Equilibrium \\
Force Equilibrium & Recognize that the condition $\Sigma$F y = 0 implies a state of force equilibrium in the vertical direction. & Force Equilibrium \\
Multiple Choice Questions & Know how to interpret and answer multiple-choice questions. & multiple choice format \\
Selecting Correct Answer & Identify the correct option based on given conditions and context. & identifying correct option \\ \hline

\end{tabular}

\label{table:oli_statics}
\end{subtable}

\begin{subtable}{\textwidth}
\centering
\caption{oli\_psychology}
\begin{tabular}{|p{3.7cm}|p{7cm}|p{3.7cm}|}
\hline
\textbf{Generated Name} & \textbf{Description} & \textbf{KC Name} \\ \hline

\multicolumn{3}{|l|}{\textbf{Item ID: 2066}} \\ \hline
Human Eye Anatomy & Understanding the different parts of the human eye and their functions. & Young-Helmholtz theory \\
Iris Location & The iris is positioned between the cornea and lens, controlling the amount of light that enters the eye. & Young-Helmholtz theory \\ \hline

\multicolumn{3}{|l|}{\textbf{Item ID: 709}} \\ \hline
Reasonable Mind Concept & Understanding that the decision involves logical planning and time management. & True or False Questions \\
Wise Mind Concept & Recognizing the balance between emotional and reasonable mindsets, though not applicable here. & Identifying emotions \\
Emotional Mind Concept & Understanding that decisions driven purely by emotions are not being considered in this scenario. & Identifying emotions \\
Time Management Skills & The ability to plan and allocate time effectively for various activities including work, study, and social events. & True or False Questions \\ \hline

\end{tabular}
\end{subtable}

\label{table:oli_statics and psychology}
\end{table*}

\begin{figure*}[h]
    \Description{Three side-by-side line charts show how AFM performance (on the y-axis, labeled RMSE) varies with the number of clusters (x-axis from 0 to 200). Each chart represents a different method: Stratified CV (left), Student CV (middle), and Item CV (right). In each chart, there are three lines for the different subject areas: Computing (blue), French (orange), and Statics (red). The trends illustrate how RMSE changes as the number of clusters increases for each subject and cross-validation approach.}
    \centering
    \includegraphics[width=0.95\linewidth, trim=0cm 0cm 0cm 0cm, clip]{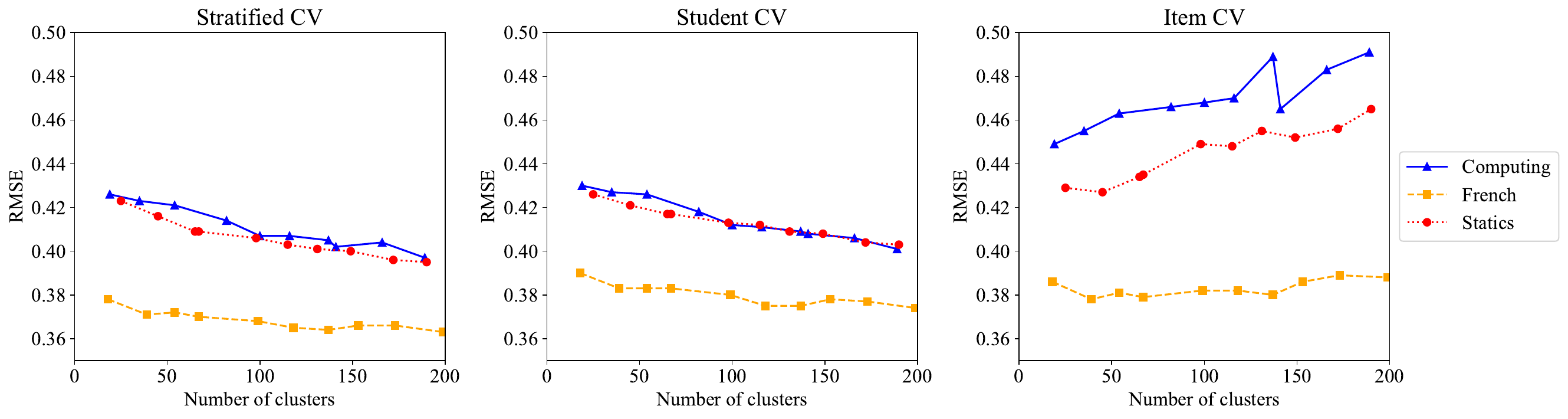}
    \caption{AFM performance change based on the number of clusters.}
    \label{fig:cluster_score}
\end{figure*}

\subsubsection{AFM Analysis}
\label{AFM}
For AFM inference, we used the Python implementation compatible with the CMU DataShop format\footnote{\url{https://github.com/cmaclell/pyAFM}}. The mapped KCs were post-processed in the same scheme as when transactions are exported with the ‘By Transaction’ option from CMU DataShop, and the inference was performed using this code. We were unable to conduct AFM analysis on the Psychology and Biology datasets because the memory usage of the code increases proportionally with the transaction length. The required memory exceeded 40GB, which could not be handled by the computational resources used in this study, specifically a MacBook M3 Pro with 36GB of shared memory. We used 3-fold cross-validation, adopting the default hyperparameters from the original AFM implementation \cite{afmslip:2015}.

\section{Results}
\label{sec:results}

\begin{table*}[t]
\centering
\caption{Summary statistics of generated KCs for different OLI courses.}
\begin{tabular}{cccc|cc}
\toprule
\textbf{Course} & \textbf{\# KCs} & \textbf{Avg. KCs per item} & \textbf{\% Multi-KC} & \textbf{\# KCs in gold} \\ 
\midrule
oli\_computing & 118 & 3.14 & 91\% & 41 \\ 
oli\_statics & 187 & 3.01 & 94\% & 81 \\ 
oli\_french & 196 & 2.91 & 94\% & 7 \\ 
oli\_biology & 197 & 2.1 & 71\% & 275 \\ 
oli\_psychology & 187 & 2.08 & 69\% & 226 \\ 
\bottomrule
\end{tabular}
\label{table:kc_summary}
\end{table*}

Figure \ref{clustered_example} provides an example of generated KCs descriptions classified into the same cluster, represented as \textit{Trigonometric Relationships}. Figure \ref{fig:cluster_score} shows the performance change of AFM with varying cluster numbers. When item information is masked, performance relies solely on the general ability of each student and KCs information, making it highly dependent on KCs quality. While overall and student-masked performance improved with more clusters, item-masked performance deteriorated, indicating that more clusters do not necessarily mean better KCs and suggesting room for improvement in verifying KCs consistency.

\begin{table}[t]
\centering
\caption{AFM scores. RMSE columns are the full cross-validation score, and - Student and - Item columns are the performance when the corresponding feature is blocked. \textit{openai} and \textit{T5-XXL} are our generated KCs, while the others were created by humans.}
\begin{tabular}[width=\textwidth]{lccc}
\toprule
Model & RMSE & - Student & - Item \\
\midrule
\textbf{OLI\_statics} \\
openai & 0.395 & \textbf{0.403} & 0.465 \\
T5-XXL & 0.395 & 0.404 & 0.478 \\
F2011 & \textbf{0.394} & \textbf{0.403} & \textbf{0.407} \\
\midrule
\textbf{OLI\_french} \\
openai & 0.363 & 0.374 & 0.388 \\
T5-XXL & 0.376 & 0.385 & 0.409 \\
Bonnie & \textbf{0.345} & \textbf{0.354} & \textbf{0.346} \\
Level4 & 0.354 & 0.358 & 0.355 \\
\midrule
\textbf{OLI\_computing} \\
openai & \textbf{0.397} & \textbf{0.401} & 0.491 \\
T5-XXL & 0.398 & 0.402 & 0.502 \\
poc\_1\_13 & 0.416 & 0.422 & \textbf{0.432} \\
poc\_1\_10 & 0.428 & 0.433 & 0.435 \\
\bottomrule
\end{tabular}
\label{fig:afm_scores}
\end{table}

Table \ref{fig:afm_scores} compares the AFM performance of KCs selected using OpenAI’s embedding API, Sentence-T5-XXL embeddings, and human experts in the \textit{statics}, \textit{french}, and \textit{computing} domains. The results using OpenAI’s embeddings consistently outperformed Sentence-T5-XXL. Given that previous research has shown the superior performance of OpenAI’s embedding models \cite{hongliurecent}, it can be concluded that as embedding models improve, the performance of KCs is likely to improve as well. However, both methods still showed comparable or better overall performance than human-created KCs, while item-blocked cross-validation performance was worse.

In the context of automatically generated KCs, the item-blocked performance of the AFM tends to be somewhat lower compared to that observed with human-defined KCs. We attribute this phenomenon to the AFM's reliance on the Opportunity Count feature. As the number of tags increases, the opportunity count values input for each prediction converge toward zero, leading to a sparsity of information that can adversely affect model performance.

As demonstrated in Table \ref{table:kc_summary}, all three subjects experienced a substantial increase in the number of generated KCs compared to the original sets. Specifically, Figure \ref{fig:cluster_score} provides compelling evidence that item-blocked performance diminishes as the number of KCs escalates. This trend is further supported by the significant disparity in AFM performance observed in the oli\_french dataset, where the original human-created KCs numbered only seven. In the computing dataset, the KC count increased from 41 to 118, and in statics from 81 to 187—approximately doubling in both cases. In contrast, the oli\_french dataset saw an increase from 7 to 196 KCs, a 28-fold amplification, which likely intensified the observed effect on AFM performance.

Thus, while the AFM is capable of effectively evaluating performance in multiple-KC systems, we infer that significant discrepancies in the number of KCs can lead to inequitable comparisons between different KC sets. This finding implies that when applying Knowledge Tracing (KT) methodologies, especially in contexts with vastly differing KC counts, it is crucial to consider the potential impact on model performance assessments. Careful examination of these factors is essential to ensure fair and accurate evaluations within educational data mining and learning analytics.

As qualitative aspects of generated KCs, Table \ref{table:oli_statics and psychology} presents randomly selected content from the oli\_statics and oli\_psych-ology datasets, showing the GPT-4o generated names and descriptions for the KCs tagged to those contents and the KC names classified by clustering. For the Biology and Psychology datasets, the higher specificity of the topics often resulted in all KCs within a single problem being classified under the same tag, explaining the relatively low Multi\_KC ratio in Table \ref{table:kc_summary}. Figure \ref{clustered_example} displays example knowledge components belonging to two clusters in oli\_statics dataset.

\subsection{Effect on Knowledge Tracing Performance}

\begin{table*}[ht]
\centering
\caption{Knowledge Tracing performance metrics (AUC). The IRT method doesn’t use any KC information. Only PFA supports multiple KCs, while the other models concatenate all KCs and treat them as a single, independent KCs.}
\begin{tabular}{clccccc}
\toprule
Knowledge Component Source & KT Model & French & Computing & Statics & Biology & Psychology \\
\midrule
\textbf{None}&IRT & 0.822 & 0.809 & 0.797 & 0.743 & 0.781 \\
\midrule
&PFA & 0.619 & 0.604 & 0.600 & 0.595 & 0.590 \\
\textbf{Random}
&DAS3H & 0.873 & 0.816 & 0.804 & 0.762 & 0.793 \\
&SAKT & 0.828 & 0.780 & 0.812 & 0.858 & 0.809 \\
&DKT & 0.925 & 0.817 & 0.860 & 0.912 & 0.822 \\
\midrule
&PFA & \textbf{0.787} & \textbf{0.723} & \textbf{0.751} & 0.666 & \textbf{0.698} \\
\textbf{LLM (Ours)}&DAS3H & 0.881 & 0.800 & 0.836 & \textbf{0.772} & \textbf{0.802} \\
&SAKT & \textbf{0.869} & \textbf{0.802} & \textbf{0.854} & 0.869 & 0.815 \\
&DKT & 0.918 & 0.835 & \textbf{0.883} & 0.915 & \textbf{0.828} \\

\midrule
&PFA & 0.752 & 0.699 & 0.693 & \textbf{0.671} & \textbf{0.698} \\
\textbf{Human}&DAS3H & \textbf{0.911} & \textbf{0.840} & \textbf{0.843} & 0.768 & 0.801 \\
&SAKT & 0.850 & 0.554 & \textbf{0.854} & \textbf{0.874} & \textbf{0.817} \\
&DKT & \textbf{0.929} & \textbf{0.868} & 0.877 & \textbf{0.918} & \textbf{0.828} \\
\bottomrule
\end{tabular}
\label{table:main}
\end{table*}

Table \ref{table:main} shows the results of the KT experiments. For PFA and DAS3H, which are logistic regression-based KT models that can utilize multiple KCs \cite{Pavlik2009PerformanceFA, Choffin2019DAS3HMS}, we find using the KCs generated by our algorithm improves performance compared to the \textbf{Random} baseline. Notably, in PFA performance, our KCs outperform those of human experts across three domains. We believe that these advantages stem from the rich information provided by the multiple KCs tags per item.

When comparing model-wise performance, our generated KCs exhibited a pattern similar to that of human expert KCs. In certain datasets and models, using LMM-generated KCs even showed a greater performance increase compared to human KCs. This, along with the previous experiments, supports the conclusion that our generated KCs explain the training data as effectively as human experts.

\section{Discussion \& Limitation}
The goal of this work was to evaluate the effectiveness of using LMMs to generate KCs directly from the text, figures, and diagrams of questions.
In an empirical evaluation using AFM, we found that the KCs generated by our LMM-based method matched the quality of human-generated knowledge components.
Our method worked across five different domains (from computing to psychology to French) and four different knowledge tracing models.
Furthermore, in models designed to work with multiple KCs per question, the KCs generated by our method outperformed human-generated KCs in four of the five datasets.
Our method has the potential to immediately improve the quality of intelligent tutoring systems by making it possible to quickly generate high quality KCs for practically any set of questions.


\subsection{Towards More Refined Domain Modeling}

As reviewed by previous work \cite{pelanek2017bayesian}, Modeling the teaching domain with KCs has the potential to achieve greater granularity and richer informational content by assigning multiple KCs to a single item, modeling hierarchical relationships among KCs, or capturing prerequisite structures. However, in practice, it has often been the case that KCs are directly derived from curricula classifications, leading to a one-to-one mapping between problems and KCs \cite{8295250}. Consequently, Transformer-based KT models, such as SAINT and SAKT, have been designed to utilize this straightforward mapping \cite{choi2020towards, pandey2019self}. However, given that real-world problems typically require the interconnection of multiple KCs, this approach deviates significantly from the intended role of KCs as \textit{``units of cognitive function"} within the original Knowledge-Learning-Instruction framework \citep{koedinger2012knowledge}.

Moreover, recent studies have increasingly diverged from the original definition of KCs, treating them as mere metadata to be leveraged for enhancing KT performance. There have been instances where KCs have been used interchangeably with terms like \textit{knowledge concepts} \cite{ai2019concept} or generalized into labels such as \textit{knowledge, tag}, or \textit{skill} \cite{8295250}. This shift risks undermining the connection between KCs and learning science fields, such as recommendation systems that rely on KCs to provide meaningful educational insights. Therefore, it is crucial to maintain the integrity of KCs, ensuring they are not reduced to simple metadata for performance enhancement purposes. We note that in this work we use "Knowledge Components" as our primary term and subsume related notions (e.g., concept, skill, tag) under this unified definition.

As large language models continue to advance, enabling the successful execution of more complex cognitive tasks, there is a growing opportunity to construct more sophisticated KC frameworks. These frameworks could consider hierarchical and prerequisite relationships with significantly reduced overhead, offering more robust and nuanced models for educational contexts. Moving beyond the AFM, there is a pressing need for enhanced methodologies that can leverage highly detailed KCs and evaluate their quality. Revisiting discussions on Cognitive Diagnosis Models may also help us remain focused on the core issues at hand \cite{de2011generalized}.

\subsection{Linking to Zero-Shot Knowledge Tracing}

A key direction for future research, as proposed by this study, is the development of Zero-Shot KT techniques mediated by KCs. Currently, KT models are predominantly optimized for interaction data, rendering assessment infeasible without prior records. This limitation is particularly problematic for ITS, where KT models a student's knowledge state based on their problem-solving history to predict future performance \cite{abdelrahman2023knowledge}. Despite the advances brought about by transformer architectures, these models still rely heavily on the statistical properties derived from problem-solving records. This approach contrasts sharply with human educators, who can intuitively assess knowledge and identify deficiencies without the need for extensive data histories.

The primary challenges associated with current KT methods include:

\begin{enumerate}
    \item The inability to manage educational content or students without prior records, leading to cold-start issues for both users and items.
    \item Biases stem from an over-reliance on statistical data, which can be influenced by the difficulty level of the content or peer interactions.
    \item In the case of Deep Neural Network models, operations that are theoretically expected are not always guaranteed. For example, even if a learner answers more questions correctly than before, the learner’s KT prediction value may be lower than it was before answering those questions \cite{kimbehavioral}.
\end{enumerate}

While human educators also face challenges with user-cold starts, they can manage biases more effectively by assessing knowledge within learning materials and identifying the essential problem-solving skills. However, the subjectivity inherent in human assessment remains a significant issue, particularly in large-scale educational systems that struggle to rely solely on human evaluation. In this context, LLMs, with their capacity to quickly comprehend content and analyze vast amounts of data, offer a promising alternative.

The most significant synergy between automatically generated KCs and Zero-Shot KT lies in the potential to enhance KC frameworks without the constant need for human expert evaluation. By leveraging performance comparisons across existing KT benchmarks, much like the automated KC evaluation platforms provided by Datashop based on the AFM \cite{koedinger2010data}, researchers can accelerate the research cycle, eliminating human bottlenecks and fostering rapid advancements in educational technology.

\subsection{Limitations}

Our study has several limitations. First, our data preprocessing led to significant losses—extraction from outdated HTML (including deprecated Flash content), potential audio-to-text conversion errors, and the exclusion of uncertain mappings reduced transaction data by more than half. Second, current KT models (SAKT and DKT) that rely on joint skill modeling are not sufficiently advanced to fully assess our detailed multi-KC approach. Finally, due to the visual nature of some problems, we relied on images rather than text, which prevented direct comparisons between Large Language Models and Large Multimodal Models.


\subsubsection{Losses in Data Preprocessing}
Refining the content data proved challenging, resulting in substantial losses. Extracting images from HTML files including deprecated Flash elements and converting audio to text introduced errors. The most substantial data loss occurred during the mapping of content data to transaction data. Due to incomplete metadata and the absence of content for some problems, we excluded any content with uncertain mapping to ensure data accuracy. This process reduced the original transaction data by over half.


\subsubsection{KT Model Limitations}

SAKT and DKT rely on joint skill modeling, which does not fully capture the nuances of our multi-KC approach. Additionally, AFM—being an older model—utilizes relatively simple feature engineering to process time-series data, limiting its ability to evaluate advanced KC details. Our primary focus was on developing detailed KCs, leaving further refinement of KT methodologies for future work. Moreover, it would be desirable for future work to support additional KT frameworks—such as the multi‐method pyKT library \cite{liupykt2022} and the LKT framework \cite{pavlik2023automated}—so that our extracted KCs can be evaluated across a broader and more diverse set of models.

\subsubsection{Access to Raw Content Data}


While we have made all generated tags and refined data publicly available, the parsed raw data remains accessible only through CMU DataShop due to their policy. Because these contents are actually used in educational settings, they cannot be made publicly available without restrictions. However, obtaining access is straightforward and promptly processed, and we have provided reproducible preprocessing code to facilitate this. We are committed to ensuring the usability of this benchmark by providing all necessary support for access.

\subsubsection{Upper Bound on the Number of Clusters}
As shown in Table~\ref{fig:cluster_score}, AFM’s overall RMSE continues to decrease even up to 200 clusters. In theory, a reversal would be expected —
 an increase in RMSE due to overfitting —
 when the cluster count becomes sufficiently large, but our experiments were restricted to at most 200 clusters. Moreover, we evaluated only three relatively small datasets (Statics, Computing, and French), which further limited our ability to explore higher cluster counts. These restrictions stem from the computational burden that grows with both the size of the dataset and the granularity of the cluster under finite resources. We anticipate that the development of more efficient algorithms for selecting the optimal number of KCs would enable exploration beyond this bound and could yield additional performance gains.

\section{Conclusion}



We have presented a novel, zero-shot approach that leverages instruction-tuned large multimodal models (LMMs) to automatically extract knowledge components (KCs) from educational multimedia content. Unlike traditional methods—which rely on human-generated labels or purely statistical techniques—our approach directly parses text, images, and audio to generate detailed KCs. Experimental evaluations across five domains and multiple knowledge tracing (KT) models (including IRT, PFA, DAS3H, SAKT, and DKT) demonstrate that the LMM-generated KCs not only match but often exceed the performance of human-defined KCs, thereby improving the accuracy and interpretability of student performance predictions.

In addition, by releasing refined KT benchmarks enriched with these automatically generated KCs, we provide a valuable resource for the community to further develop advanced KT methodologies. While our findings highlight the promise of automated KC extraction in enhancing intelligent tutoring systems, they also reveal key limitations—such as significant data losses during preprocessing and the constraints of existing KT models in fully capturing the nuances of multi-KC assignments—that must be addressed in future research. Moving forward, integrating more sophisticated domain modeling techniques and exploring zero-shot KT strategies will be crucial for developing more personalized and scalable educational systems.

Overall, our work lays a strong foundation for the next generation of content-aware KT models, bridging the gap between modern AI capabilities and educational practice.
\balancecolumns
\newpage
\bibliographystyle{abbrv}
\bibliography{sigproc}



\end{document}